%% file: main.tex
\documentclass[runningheads]{llncs}

\usepackage{eccv}

\input{preamble}

\usepackage{eccvabbrv}

\usepackage{graphicx}
\usepackage{booktabs}

\usepackage[accsupp]{axessibility}
\usepackage{hyperref}
\usepackage{orcidlink}

\begin{document}

\title{The Devil is in the Statistics: Mitigating and Exploiting Statistics Difference for Generalizable Semi-supervised Medical Image Segmentation}
\titlerunning{The Devil is in the Statistics}

\author{
Muyang Qiu\inst{1,2},
Jian Zhang\inst{1,2},
Lei Qi\inst{3}, \\
Qian Yu\inst{4},
Yinghuan Shi\inst{1,2}\thanks{Corresponding Author.},
Yang Gao\inst{1,2}}
\authorrunning{Qiu et al.}

\institute{State Key Laboratory for Novel Software Technology, Nanjing University \and
National Institute of Healthcare Data Science, Nanjing University \\
\email{\{qmy,zhangjian7369\}@smail.nju.edu.cn, \{syh,gaoy\}@nju.edu.cn} \and
School of Computer Science and Engineering, Southeast University \\
\email{qilei@seu.edu.cn} \and
School of Data and Computer Science, Shandong Women’s University \\
\email{yuqian@sdwu.edu.cn}}

\maketitle
\input{sec/0_abstract}
\input{sec/1_intro}

\input{sec/2_rel}
\input{sec/3_method}
\input{sec/4_exp}
\input{sec/5_cl}
\section*{Acknowledgements}
\sloppy  
 This work is supported by the NSFC Project (62222604, 62206052, 62192783), Jiangsu Natural Science Foundation Project (BK20210224), China Postdoctoral Science Foundation (2024M750424), the Fundamental Research Funds for the Central Universities (020214380120), the State Key Laboratory Funds for Key Project (ZZKT2024A14), and Shandong Natural Science Foundation (ZR2023MF037).

%
%
\bibliographystyle{splncs04}
\bibliography{main}

\ifx\excludeSupplementary\undefined 
    \newpage
    \input{sec/X_suppl}
\fi

\end{document}

%% file: preamble.tex
\usepackage[dvipsnames]{xcolor}
\usepackage{soul}

\usepackage{indentfirst}
\usepackage{bbm}
\usepackage[export]{adjustbox}

\usepackage{multirow}
\usepackage{colortbl}
\definecolor{mygray}{gray}{.9}
\definecolor{mygreen}{RGB}{53, 160, 20}

\newcommand{\pub}[1]{\ignorespaces}
\newcommand{\stdev}[1]{\tiny $\pm$#1}

\usepackage{algorithm}
\usepackage{listings}

\usepackage{wrapfig}

%% file: sec/0_abstract.tex
\begin{abstract}
Despite the recent success of domain generalization in medical image segmentation, voxel-wise annotation for all source domains remains a huge burden.
Semi-supervised domain generalization has been proposed very recently to combat this challenge by leveraging limited labeled data along with abundant unlabeled data collected from multiple medical institutions, depending on precisely harnessing unlabeled data while improving generalization simultaneously.
In this work, we observe that domain shifts between medical institutions cause disparate feature statistics, which significantly deteriorates pseudo-label quality due to an unexpected normalization process. Nevertheless, this phenomenon could be exploited to facilitate unseen domain generalization.
Therefore, we propose 1) multiple statistics-individual branches to mitigate the impact of domain shifts for reliable pseudo-labels and 2) one statistics-aggregated branch for domain-invariant feature learning. Furthermore, to simulate unseen domains with statistics difference, we approach this from two aspects, \ie, a perturbation with histogram matching at image level and a random batch normalization selection strategy at feature level, producing diverse statistics to expand the training distribution.
Evaluation results on three medical image datasets demonstrate the effectiveness of our method compared with recent SOTA methods.
The code is available at \url{https://github.com/qiumuyang/SIAB}.

\keywords{Medical image segmentation \and Semi-supervised domain generalization}
\end{abstract}

%% file: sec/1_intro.tex
\section{Introduction}
\label{sec:intro}

Medical image segmentation has witnessed substantial advancements recently, driven by the advent of deep convolutional neural networks \cite{ronneberger2015unet, milletari2016vnet, chen2021transunet}. 
However, the efficacy of existing methods can be impeded by the distribution shift between training and test datasets due to inherent differences in imaging protocols, modalities, and patient populations across medical institutions, which is frequently encountered in real-world clinical scenarios.
Therefore, domain generalization (DG) has been proposed to tackle this challenge \cite{wang2022generalizing, guo2023domaindrop}, aiming to train models that are generalizable to unseen domains only using data collected from multiple relevant source domains.

Although existing DG methods present practical solutions to combat domain shifts, they often require extensive annotations for each source domain. This is particularly expensive in medical image segmentation due to the need for precise delineation and expert knowledge \cite{tajbakhsh2020embracing}. 
Semi-supervised domain generalization (SSDG) \cite{zhou2021stylematch, liu2021semimeta}, harmonizing the strengths of semi-supervised learning (SSL) \cite{cheplygina2019not} and DG, has emerged as a promising new direction to alleviate this annotation burden.
It extends the capabilities of conventional DG by leveraging the abundant but unlabeled information within each source domain.

\begin{figure}[t]
\centering
\subfloat[Domain shifts]{%
    \label{fig:fundus-sample}%
    \includegraphics[width=0.31\columnwidth,valign=t]{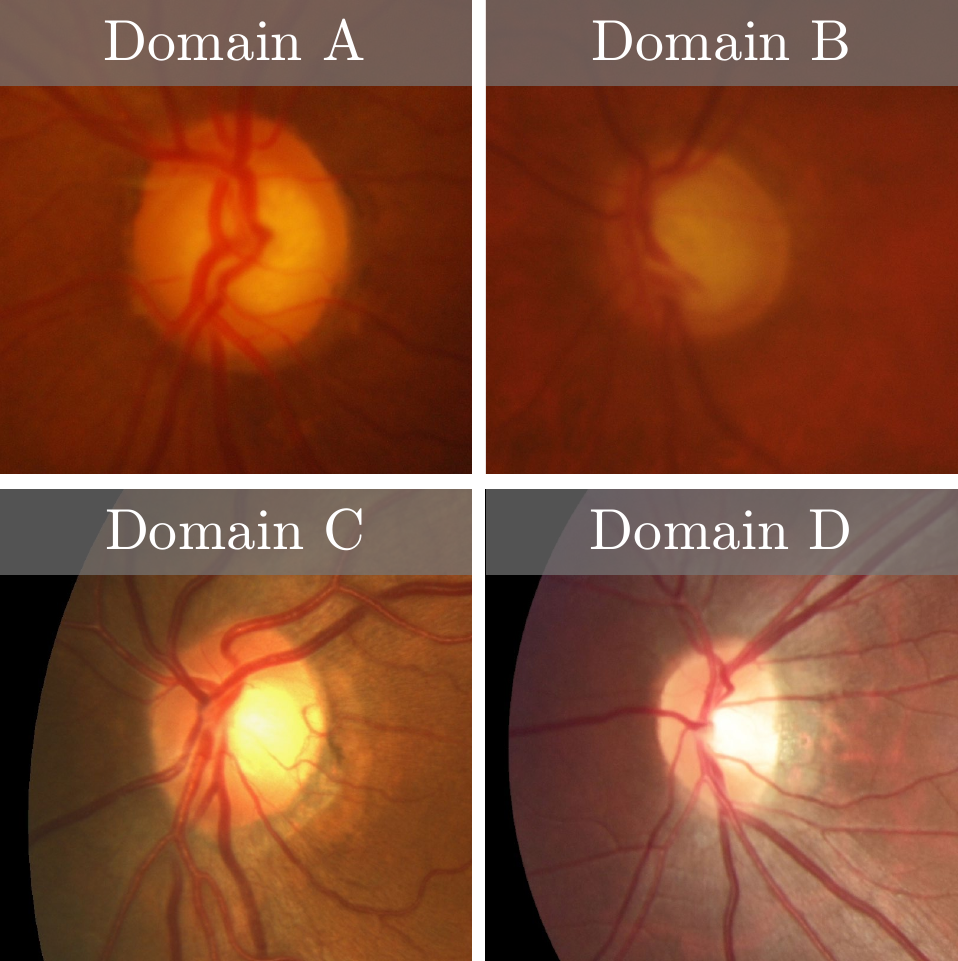}%
    \vphantom{\includegraphics[width=0.355\columnwidth,valign=t]{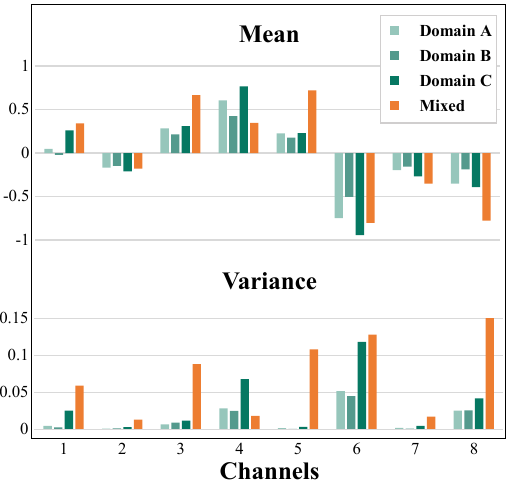}}%
}%
\hfill
\subfloat[Feature Statistics]{%
    \label{fig:statistics}%
    \includegraphics[width=0.355\columnwidth,valign=t]{fig/intro-stat.pdf}%
}%
\hfill
\subfloat[Pseudo-label]{%
    \label{fig:pseudo}%
    \includegraphics[width=0.239\columnwidth,valign=t]{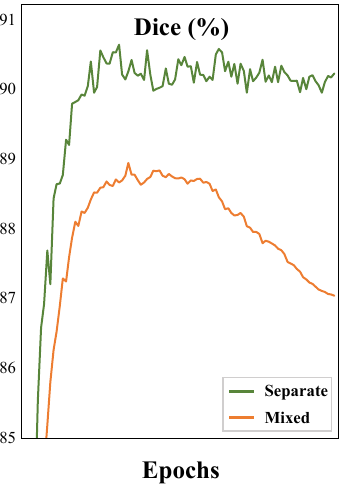}%
}
\caption{(a) Images sampled from 4 domains of the Fundus dataset. (b) Differences in feature statistics captured by batch normalization layers on the Fundus dataset. (c) Comparison of pseudo-label quality (measured by dice similarity coefficient against ground truth) between separate and mixed domain feature normalization.}
\label{fig:intro}
\end{figure}

Typically, applying well-established SSL or DG medical image segmentation methods to SSDG exposes limitations arising from the unique challenges posed by SSDG. 
DG methods, although focusing on tackling domain shifts and enhancing generalization, usually rely on full label information \cite{wang2020dofe, liu2021feddg} or indirectly utilize unlabeled data through self-supervised tasks \cite{zhou2022ramdsir}.
SSL methods, while specialized in exploiting unlabeled data, are not explicitly designed to handle the complexities of multiple domains and may suffer from the distribution shift \cite{bai2023bidirectional}.
To tackle the aforementioned issues, the main goals of SSDG are twofold: \emph{improving the quality of pseudo-labels under the domain shift} and simultaneously \emph{promoting generalization ability utilizing all available data}.
Prior works for SSDG are primarily built upon SSL methods \cite{zhou2021stylematch, liu2021semimeta, yao2022cacps}, guided by the belief that a more robust decision boundary can be learned by exploiting all available data.
For instance, Yao~\etal~\cite{yao2022cacps} incorporated a Fourier transform-based confidence-aware module into CPS~\cite{chen2021cps}, filtering high-confidence pixels to obtain more reliable pseudo supervision signals for unlabeled data in medical image segmentation.
Despite their success, the enhancement of pseudo-labels is achieved through post-processing of the model's outputs, refining these predictions as a secondary step, which limits the quality of the initial results. This raises a crucial question: \emph{Is there a more direct and graceful approach to seamlessly mitigate the impact of domain shifts throughout the prediction process?}

It is well known that batch normalization (BN) plays an important role throughout deep models by accelerating convergence and stabilizing training \cite{ioffe2015batch}. Its function lies in normalizing input features via batch statistics, \ie, mean and variance.
However, our observation reveals a significant challenge when mini-batches are drawn from diverse domains: \textbf{domain shifts induce substantial differences in feature statistics} within these batches, as depicted in \cref{fig:fundus-sample,fig:statistics}. 
For multi-site medical images, domain shifts mainly manifest as grayscale differences. While these differences might be imperceptible to the human eye, they are prominently captured by the BN.
Adopting one single BN for all source domains disregards these differences, resulting in a mismatch between the normalization statistics optimal for each domain and the statistics actually used.
Consequently, it adversely affects the normalized features, leading to a negative impact on pseudo-labels, as in \cref{fig:pseudo}.  

Motivated by this, we propose statistics-individual branches (SIBs) to separately normalize image features for each medical institution, alleviating the degradation of pseudo-label quality caused by the undesired normalization process.
We equip each SIB with dedicated BNs, thereby enabling independent capture and adaptation to the statistical nuances of features from each distinct domain.
This design directly diminishes the impact of domain shifts during prediction, minimizing the need for additional output processing.

Besides, we introduce a statistics-aggregated branch (SAB) as a complement to SIBs.
Trained on data from all medical institutions, SAB is tasked with generating predictions consistent with those of SIBs, aimed at learning domain-invariant features and contributing to improved generalization performance.

While statistics discrepancies can be detrimental to pseudo-labels, we identify an opportunity to boost overall generalization by artificially simulating domain shifts during the training stage.
Therefore, we further introduce additional perturbations to create statistical variations.
SAB is then required to maintain consistency despite both suboptimal normalization statistics aggregated from multiple domains and these perturbations, facilitating robust feature learning.
We integrate histogram matching at the image level for grayscale distribution variations. Leveraging SIBs, we develop a feature-level perturbation method utilizing affine parameters in various BNs. These parameters, adjusting the mean and variance of features, implicitly model the unique feature distribution of their respective domains, hence ideal for introducing targeted perturbations.

Our contributions can be summarized as follows:

\begin{itemize}

    \item Based on the pros and cons of domain statistics difference, we propose a multi-branch framework for SSDG medical image segmentation, utilizing both statistics-individual and statistics-aggregated branches to collaboratively preserve pseudo-labels and improve generalization.

    \item We propose additional multi-level perturbation streams at both the image and feature levels to simulate domain shifts, expanding the training distribution to achieve better robustness for the statistics-aggregated branch.

    \item We demonstrate the effectiveness of our method on three medical image segmentation benchmarks by comparison with several classic and state-of-the-art methods. 
    
\end{itemize}

%% file: sec/2_rel.tex
\section{Related Work}
\label{sec:rel}

\textbf{Semi-supervised Medical Image Segmentation.} 
Existing methods for semi-supervised medical image segmentation can be broadly categorized into two main streams: self-training \cite{lee2013pseudo, bai2017semi, shi2021inconsistency} and consistency regularization \cite{cai2023orthogonal, luo2021semi, yu2019uncertainty}.
Self-training methods involve iterative training by generating pseudo-labels or using other techniques. 
For instance, Wang~\etal \cite{wang2023mcf} developed a mutual correction framework to explore network bias correction for pseudo-labels.
Consistency regularization-based methods generate diverse predictions through input or feature perturbations, multiple tasks, or network structure, and enforce consistency among them.
Li~\etal \cite{li2020transformation} perturbed inputs using various transforms to encourage consistency between the student and teacher model.
Luo~\etal \cite{luo2021semi} established dual tasks for segmentation and level-set regression. Wu~\etal \cite{wu2021semi} integrated an extra decoder to the segmentation network to obtain discrepancies. 
Bai~\etal \cite{bai2023bidirectional} highlighted the distribution gap between labeled and unlabeled data, tackling it with a bidirectional copy-paste strategy.
Most semi-supervised methods neglect distribution differences between domains, risking performance degradation under domain shifts. However, our method takes this into account and minimizes the impact of domain shifts during pseudo-label generation.

\textbf{Domain Generalization.}
Extensive efforts have been made in the field of DG. Several methods \cite{su2023rethinking, lyu2022aadg} incorporate data augmentation techniques to broaden the data distribution and enhance the robustness of models.
Additionally, advances in episodic training strategy have driven the emergence of meta-learning-based methods \cite{li2018learning, dou2019domain, liu2020saml, wang2023generalizable}. Liu \etal \cite{liu2021feddg} utilized this strategy with continuous frequency space interpolation in a federated DG scenario.
Learning domain-invariant features by minimizing the domain discrepancy across multiple source domains is another common approach \cite{wang2020dofe}.
Another category of DG methods constructs self-supervised tasks for regularization. 
Zhou~\etal \cite{zhou2022ramdsir} introduced an image restoration task after distortion in the frequency domain.
While self-supervised tasks offer certain advantages in SSDG by leveraging unlabeled data, DG methods still fall behind in harnessing unlabeled data reliably.
Besides, several methods employ additional network structures such as convolution and normalization layers. Hu~\etal \cite{hu2022domain} tackled DG using a dynamic convolution with parameters conditioned on predicted domain code and global features.  Seo~\etal \cite{seo2020learning} incorporated a mixture of batch and instance normalization for each domain and averaged their outputs as the prediction for the unseen domain.
In contrast, our design introduces a different structure, including both individual batch normalization layers to enhance pseudo-labels and an inter-domain shared layer for generalization, which is aligned with the two major objectives of SSDG.

%% file: sec/3_method.tex
\begin{figure*}[t]
    \centering
    \includegraphics[width=\linewidth]{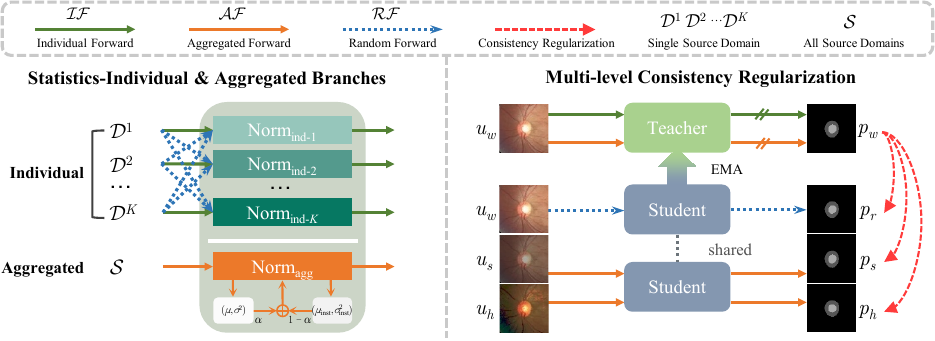}
    \caption{Overview of our method. The left figure illustrates the structure of the statistics-individual and aggregated branches. The right figure illustrates the consistency regularization framework extended with diverse perturbations. $u_w, u_s, u_h$ denote the weak, strong, and histogram matching-augmented unlabeled images respectively, and $p_w, p_s, p_h$ are the corresponding predictions. $p_r$ is the prediction of $u_w$ through random forward. Arrows symbolize the forward processes based on the proposed branches during prediction. Dashed lines signify that one branch is randomly selected each time. $\bigoplus$ denotes element-wise addition.}
    \label{fig:method}
\end{figure*}

\section{Method}
\label{sec:method}

We start by presenting the preliminaries, followed by the introduction of our method in \cref{sec:siab} and \cref{sec:cons}. Finally, we briefly discuss the complexity of our method in \cref{sec:com-dis}. \cref{fig:method} provides an overview of the method.

\subsection{Preliminaries}
\label{sec:prelim}

We denote the training set consisting of $K$ source domains as $\mathcal{S} = \{ \mathcal{D}^{1}, \mathcal{D}^{2}, $ \\ $\cdots, \mathcal{D}^{K} \}$. 
Each source domain consists of the labeled part $\mathcal{D}^{k}_{l} = \{ (x^{k}_{i}, y^{k}_{i}) \}^{N^{k}_{l}}_{i=1}$ and the unlabeled part $\mathcal{D}^{k}_{u} = \{ u^{k}_{i} \}^{N^{k}_{u}}_{i=1}$, where $x^{k}_{i}, y^{k}_{i}, u^{k}_{i}$ are the $i$-th labeled image, segmentation map and unlabeled image, respectively. $N^{k}_{l}$ and $N^{k}_{u}$ are the number of labeled and unlabeled samples in the $k$-th domain.
We expect to train a model generalizable to an unseen test domain, exploiting the unlabeled images with limited labeled data in the presence of domain shifts. 

We first set up a mean-teacher-based \cite{tarvainen2017meanteacher} semi-supervised baseline method for the leverage of unlabeled data. 
The overall objective function for the baseline can be formulated as
\begin{equation}
    \mathcal{L} = \mathcal{L}_{x} + \mathcal{L}_{u},
\end{equation}
where $\mathcal{L}_{x}$ is the supervised loss, usually a combination of cross-entropy loss and Dice loss \cite{milletari2016vnet} for medical image segmentation, and $\mathcal{L}_{u}$ is the unsupervised loss.
We embrace the ``weak-to-strong'' consistency regularization \cite{sohn2020fixmatch, yang2023revisiting} for the unsupervised loss, given that strong augmentation changes the image appearance (in brightness, contrast, saturation, \etc) by color jittering, aligning well with the manifestation of domain shifts in multi-site medical images.

To be specific, we first randomly sample a pair of weak and strong augmentation, denoted as $\mathcal{A}_{w}$ and $\mathcal{A}_{s}$, respectively. Subsequently, utilizing the teacher model $\hat{f}$
which is updated through exponential moving average (EMA) using the student's weights, 
we generate predictions for the weak-augmented unlabeled image $\mathcal{A}_{w}(u)$. Finally, we guide the student model $f$ to predict the strong-augmented counterpart $\mathcal{A}_{s}(u)$ with the pseudo-label obtained from the teacher model, by enforcing the cross-entropy consistency loss $\mathrm{H}\left(\cdot, \cdot\right)$ between them:
\begin{equation}
    \label{eq:unlabeled}
    \mathcal{L}_{u} = \mathrm{H}\left(\hat{f}\left(\mathcal{A}_{w}(u)\right), f\left(\mathcal{A}_{s}(u)\right)\right).
\end{equation}
In practice, we use one-hot pseudo-label and confidence thresholding:
\begin{equation}
    \label{eq:unlabeled-full}
    \begin{aligned}
    \mathcal{L}_{u} & = \mathbbm{1}\left(\max\left(p_{w}\right) \geqslant \tau \right) \mathrm{H}\left(\hat{p}_{w}, f\left(\mathcal{A}_{s}(u)\right)\right), \\
    p_{w} & = \hat{f}\left(\mathcal{A}_{w}(u)\right), \quad \hat{p}_{w} = \mathop{\mathrm{argmax}}(p_{w}),\\
    \end{aligned}
\end{equation}
where $\mathbbm{1}(\cdot)$ is the indicator function, $\tau$ denotes the confidence threshold, $p_w$ is the prediction of the weakly augmented unlabeled image, and $\hat{p}_{w}$ is the one-hot pseudo-label, respectively.
We will omit the indicator function term and abbreviate it as the form of \cref{eq:unlabeled} in the following sections.

\subsection{Statistics-individual and Statistics-aggregated Branches}
\label{sec:siab}

\textbf{Statistics-individual branches for pseudo-labeling.}
To alleviate the impact of domain shifts on pseudo-labels, we propose the incorporation of statistics-individual branches. Each branch includes dedicated BN layers throughout the forward propagation process to capture domain statistics individually while sharing the remaining network parameters across all branches.
For the batch activation $\mathbf{x}_{d} \in \mathbb{R}^{N \times C \times H \times W}$ from domain $d$, where $N, C, H, W$ are the batch size, number of channels, height and width of the feature map, respectively, BNs in the $d$-th SIB normalizes it by channel-wise mean $\mu_{d}$ and variance $\sigma^2_{d}$:
\begin{equation}
    \label{eq:norm_ind}
    \mathrm{Norm}_{\mathrm{ind}}(\mathbf{x}_{d}) = \gamma_{d} \frac{\mathbf{x}_{d} - \mu_{d}}{\sqrt{\sigma_{d}^2 + \epsilon}} + \beta_{d},
\end{equation}
where $\gamma_{d}$ and $\beta_{d}$ are the affine parameters, $\epsilon$ is a small positive constant to avoid division by zero, and the mean and variance of channel $c$ are calculated by:
\begin{equation}
\begin{aligned}
    \mu_{d,c} &= \frac{1}{NHW} \sum\limits_{n=1}^{N} \sum\limits_{h=1}^{H} \sum\limits_{w=1}^{W} \mathbf{x}_{d}[n, c, h, w], \\
    \sigma_{d, c}^2 &= \frac{1}{NHW} \sum\limits_{n=1}^{N} \sum\limits_{h=1}^{H} \sum\limits_{w=1}^{W} \Big(\mathbf{x}_{d}[n, c, h, w] - \mu_{d,c}\Big)^2. \\
\end{aligned}    
\end{equation}

We exclusively forward the weak-augmented images through SIBs, as weak augmentations (mainly geometric transformations) minimally affect the style and statistical properties of input images. Thus, images within each domain maintain a relatively consistent distribution without disturbing the individual branches.
Conversely, strong augmentations introduce significant statistical variations as discussed in \cite{yuan2021simple}, making the augmented images unsuitable for the SIBs.

\textbf{Statistics-aggregated branch for generalization.}
We then introduce a SAB to complement the SIBs. Considering the complexity of the input distribution it handles, which includes samples from various domains and those affected by strong augmentation, we leverage both batch and instance statistics for normalization. This approach suppresses instance-specific style information with instance statistics while preserving inter-instance discriminability with batch statistics. We adopt a linear combination of them \cite{nam2018batch, seo2020learning} as the final normalization statistics in the SAB.
Denoting the instance mean and variance as $\mu_{\mathrm{inst}}$ and $\sigma_{\mathrm{inst}}^2$, the normalization process can be formulated as:
\begin{equation}
\begin{gathered}
\label{eq:inv}
    \mathrm{Norm}_{\mathrm{agg}}(\mathbf{x}_{d}) = \gamma_{\,\mathrm{agg}} \frac{\mathbf{x}_{d} - \mu_{\mathrm{agg}}}{\sqrt{\sigma_{\mathrm{agg}}^2 + \epsilon}} + \beta_{\mathrm{agg}}, \\
    \mu_{\mathrm{agg}} = \alpha \odot \mu_{d} + (1 - \alpha) \odot \mu_{\mathrm{inst}}, \\
    \sigma^2_{\mathrm{agg}} = \alpha \odot \sigma^2_{d} + (1 - \alpha) \odot \sigma^2_{\mathrm{inst}}, \\
\end{gathered}
\end{equation}
where $\gamma_{\,\mathrm{agg}}$ and $\beta_{\mathrm{agg}}$ are the affine parameters. $\odot$ means element-wise multiplication and $\alpha \in [0, 1]^C$ is the learnable mixing coefficient for each channel.

\textbf{Consistency learning.}
We denote predicting the sample using its respective SIB as individual-forward ($\mathcal{IF}$) and using SAB as aggregated-forward ($\mathcal{AF}$). We then denote $m^\mathcal{F}(x)$ as the output of applying the forward strategy $\mathcal{F}$ to model $m$ with input $x$. The procedure of weak-to-strong consistency learning can be improved with the proposed statistics-individual and aggregated branches as follows. We first acquire predictions for weak-augmented unlabeled images using $\mathcal{IF}$ and $\mathcal{AF}$ via the teacher model $\hat{f}$:
\begin{equation}
\begin{gathered}
    p_w^{\mathcal{IF}} = \hat{f}^{\mathcal{IF}}(\mathcal{A}_{w}(u)), \quad
    p_w^{\mathcal{AF}} = \hat{f}^{\mathcal{AF}}(\mathcal{A}_{w}(u)).
\end{gathered}
\end{equation}
We then ensemble them for better pseudo-labels:
\begin{equation}
    p_w = \left(p_w^{\mathcal{IF}} + p_w^{\mathcal{AF}}\right) / 2,
    \quad
    \hat{p}_w = \mathop{\mathrm{argmax}}(p_{w}).
\end{equation}
Finally, we encourage a consistent prediction of strong-augmented images from SAB of the student model $f$ in alignment with the pseudo-label:
\begin{equation}
    \label{eq:strong}
    p_s = f^{\mathcal{AF}}(\mathcal{A}_{s}(u)), \quad
    \mathcal{L}_{s} = \mathrm{H}(\hat{p}_w, p_s).
\end{equation}

\textbf{Supervised loss.} We train both statistics-individual and aggregated branches with cross-entropy and Dice loss using weak-augmented labeled images to ensure stable and accurate statistics information.

\begin{figure}[t]
    \centering
    \begin{minipage}{0.52\textwidth}
        \centering
        \includegraphics[width=\linewidth,valign=c]{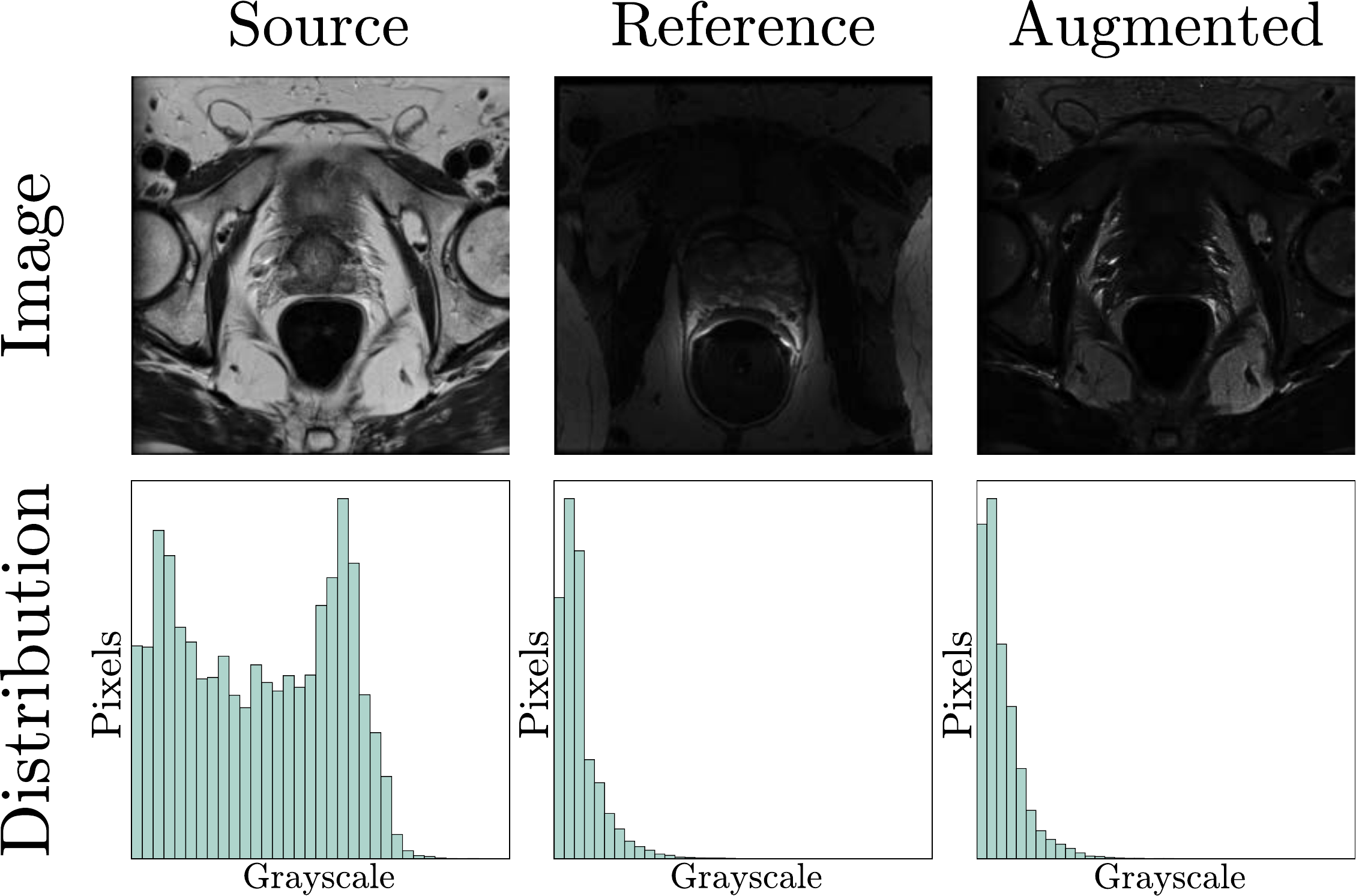}%
        \vphantom{\includegraphics[width=0.69\linewidth,valign=c]{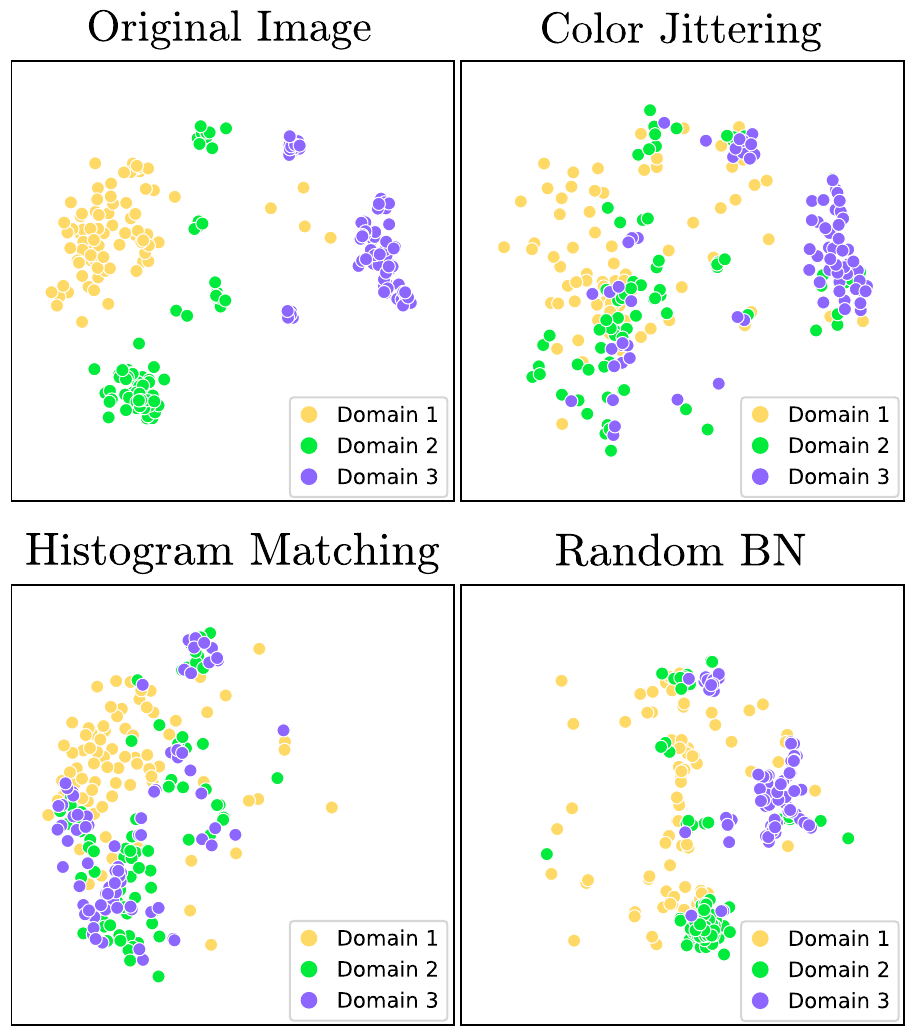}}%
        \caption{Illustration of histogram matching. The top row represents the images from the Prostate dataset. The bottom row visualizes their grayscale distribution.}
        \label{fig:hist}
    \end{minipage}\hfill
    \begin{minipage}{0.45\textwidth}
        \centering
        \includegraphics[width=0.8\linewidth]{fig/tsne.pdf}
        \caption{T-SNE visualization of image features after applying different perturbations. Different colors represent different domains. Best viewed in color.}
        \label{fig:tsne}
    \end{minipage}
\end{figure}

\subsection{Multi-level Consistency Regularization}
\label{sec:cons}

To facilitate unseen domain generalization, we aim to fully exploit the potential of statistics differences, simulating various distributions by introducing multi-level perturbations and consistency regularization.

\textbf{Strong and style augmentation at the image level.}
We have previously discussed strong augmentations, which involve random appearance perturbations to diversify training data distribution and assist the model in adapting to domain shifts. However, an excessive level of perturbation resulting from randomness can compromise the semantic integrity of images, impeding effective semantic learning.
To overcome this limitation, we leverage the diverse distribution information from existing source domains for a controllable perturbation.
Specifically, we employ histogram matching \cite{pizer1987adaptive} to align the grayscale distribution between images, as illustrated in \cref{fig:hist}.
We then introduce a new regularization stream named ``weak-to-style'' into the consistency regularization framework.
By exposing the model to images with similar semantics but diverse styles across multiple domains, it is anticipated that the model will learn domain-invariant features, robust against various perturbations introduced by both color jittering and histogram matching.
Furthermore, while color jittering diversifies source domain distribution through randomness, histogram matching bridges the distribution gap between source domains by leveraging known information. 
Both augmentation strategies synergize effectively, forming a complementary approach to enhancing model robustness.
Denoting the histogram matching operation as $\mathcal{A}_{h}$, the loss function of weak-to-style consistency regularization is similar to \cref{eq:strong}:
\begin{equation}
    p_h = f^{\mathcal{AF}}(\mathcal{A}_{h}(u)), \quad
    \mathcal{L}_{h} = \mathrm{H}(\hat{p}_w, p_h).
\end{equation}

\textbf{Normalization-selection-based perturbation at the feature level.}
Despite style diversity being improved by strong and style augmentations, both consistency regularization streams are confined to the image level, hindering the model from exploring a wider range of perturbations.

To this end, we propose to further leverage SIBs for feature-level statistics perturbation. Affine parameters within distinct BN layers inherently encapsulate variations in features across diverse domains. Hence, opting for the BN layer and utilizing its affine parameters to introduce feature perturbations becomes a reasonable strategy. To be specific, for the batch activation $\mathbf{x}_{d}$ from domain $d$ which should have been normalized by the BNs in the $d$-th SIB, we randomly select another branch $d'$, using its affine parameters
to substitute the ones in \cref{eq:norm_ind}
as the source of perturbation with a probability of $p$, which controls the intensity of the perturbation:
\begin{equation}
\begin{gathered}
    \mathrm{Norm}_{\mathrm{rand}}(\mathbf{x}_{d}) = \gamma_{d'} \frac{\mathbf{x}_{d} - \mu_{d}}{\sqrt{\sigma_{d}^2 + \epsilon}} + \beta_{d'}, \\ 
    d' \ne d, d' \in \{1, 2, \cdots, K\}.
\end{gathered}
\end{equation}
We stop the gradient for BN affine parameters during backward propagation, regardless of whether the perturbation is activated, to prevent interference with SIBs.

The feature perturbed forward process with the random BN selection strategy is denoted as random-forward ($\mathcal{RF}$). Its corresponding loss $\mathcal{L}_r$, is defined as:
\begin{equation}
     p_r = f^{\mathcal{RF}}(\mathcal{A}_{w}(u)), \quad
    \mathcal{L}_{r} = \mathrm{H}(\hat{p}_w, p_r).
\end{equation}

\cref{fig:tsne} illustrates the t-SNE visualization results of the image features extracted from the Fundus dataset. It showcases the effects of various perturbations in introducing statistics discrepancies and simulating domain shifts.

In summary, the unsupervised loss contains three terms, each representing one stream of consistency regularization:
\begin{equation}
    \label{eq:overall_ulb}
    \mathcal{L}_{u} = \mathcal{L}_{s} + \lambda_{h} \mathcal{L}_{h} + \lambda_{r} \mathcal{L}_{r},
\end{equation}
where $\lambda_{h}, \lambda_{r}$ balance the weight between different consistency streams.

\subsection{Complexity Discussion}
\label{sec:com-dis}
The primary purpose of the SIBs during training is to provide accurate pseudo-labels for SAB. These branches are tailored to their respective domains and thus may not generalize well to unseen distributions. Consequently, during the inference stage, we discard them to attain a parameter size similar to that of a conventional model structure.
Quantitative results are reported in \cref{sec:ablation}.

%% file: sec/4_exp.tex
\input{tbl/prostate}
\begin{figure}[t]
   \centering
   \includegraphics[width=\linewidth]{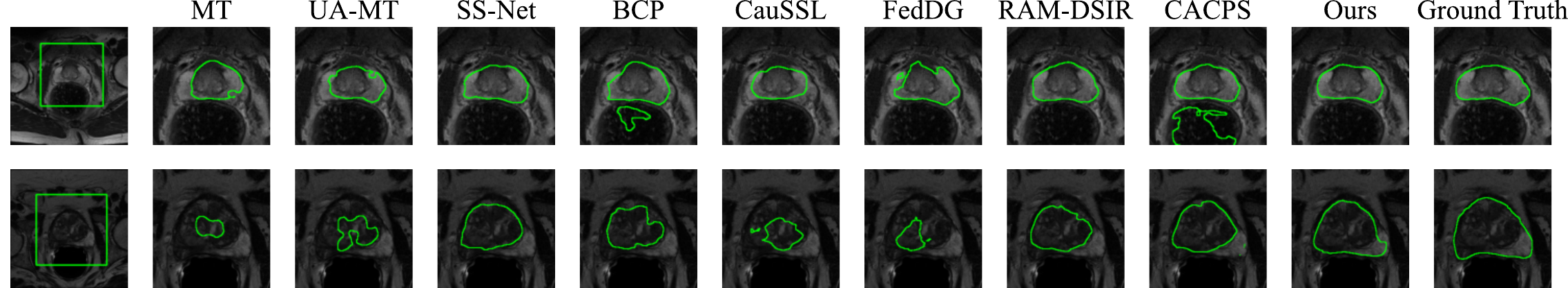}
   \caption{Visualization results from the Prostate dataset.}
   \label{fig:prostate}
\end{figure}

\section{Experiment}
\label{sec:exp}

\subsection{Datasets}

We evaluate our method on three datasets, namely, Prostate \cite{liu2020saml}, Fundus \cite{wang2020dofe}, and  M\&Ms \cite{campello2021multi}. The first two datasets are commonly used in conventional DG medical image segmentation tasks \cite{wang2020dofe,liu2021feddg,zhou2022ramdsir}, while the other in SSDG \cite{liu2021semimeta,yao2022cacps}.

\textbf{Prostate} dataset collects prostate T2-weighted MRI images from 6 medical centers for prostate segmentation. The original 3D volumes are sliced in the axial plane and resized to $288 \times 288$ for training. At test time, slices not containing any prostate region are discarded, following \cite{liu2020saml,zhou2022ramdsir}.

\textbf{Fundus} dataset collects retinal fundus images from 4 medical centers for optic cup and disc (OC and OD) segmentation. The original resolution of images is $800 \times 800$ and we resize them to $256 \times 256$ for memory efficiency at train time. At test time, the input images are downsampled and the predicted masks are then upsampled for evaluation.

\textbf{M\&Ms} dataset collects MRI cardiac images from 4 magnetic resonance scanner vendors. Each subject is presented in four dimensions (3D volumes with phase information) and only the end-systole with the end-diastole phases are annotated. Each volume is sliced, center-cropped, and then resized to $288 \times 288$ as the network input. We follow \cite{yao2022cacps} to split the labeled and unlabeled data.

\subsection{Experimental Setting}

Following the established convention in DG, we apply the leave-one-domain-out strategy to evaluate the performance of a method, \ie, choose one domain as the unseen test domain and the rest for training. For each training source domain, we use the same unlabeled split consistently among different methods.

\textbf{Implementation details.} We employ U-net \cite{ronneberger2015unet} as the basic segmentation model and implement it with PyTorch on two NVIDIA GeForce RTX 3090 GPUs. The training iterations for each dataset are $2 \times 10^4, 2 \times 10^4, 4 \times 10^4$, and the batch sizes are $12, 10, 18$, respectively. We adopt AdamW \cite{Loshchilov2019adamw} with a learning rate of $1 \times 10^{-3}$ and a weight decay of 0.01 as the optimizer. We set the parameters $\lambda_{h}$ and $\lambda_{r}$ in \cref{eq:overall_ulb} as 1.0 and 0.2, respectively. 
Weak augmentation employed in our consistency regularization framework consists of random rotation and flip, and strong augmentation includes color jittering and blur.

\textbf{Evaluation metrics.} We adopt two metrics for evaluation. Dice similarity coefficient (Dice) measures the overlap between two binary masks. Average surface distance (ASD) measures the average distance between the surfaces of two segmented objects. Higher Dice or lower ASD indicates better performance.

\input{tbl/fundus}
\begin{figure}[t]
   \centering
   \includegraphics[width=\linewidth]{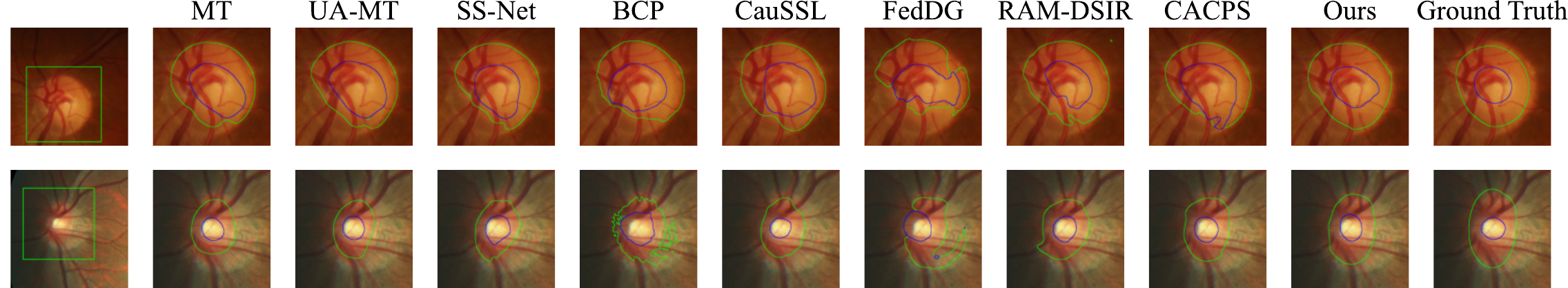}
   \caption{Visualization results from the Fundus dataset. \textcolor{blue}{Blue} contours for optic cup and \textcolor{green}{green} for optic disc.}
   \label{fig:fundus}
\end{figure}

\subsection{Comparison with State-of-the-Art Methods}

\textbf{Compared methods.}
We compare our method with classic and state-of-the-art (SOTA) SSL, DG, and SSDG methods for medical image segmentation. SSL methods utilize both labeled and unlabeled data from multiple domains without distinction. DG methods, typically for fully supervised scenarios, do not utilize unlabeled data unless incorporating self-supervised tasks.

\textbf{Results on the Prostate dataset.} 
We allocate 30\% labeled data and 70\% unlabeled for each domain in the Prostate dataset. The results are listed in \cref{table: prostate}. SSL methods exhibit limited performance on challenging domains, particularly domain E. DG methods, while effective on some domains, cannot fully leverage unlabeled data to achieve competitive overall results. Our method achieves the best Dice and ASD, with an improvement of 8.00\% and 9.61\% in terms of average Dice compared to the best-performing SSL and DG methods, respectively. Additionally, we outperform CACPS, the previous SOTA in SSDG, by 2.26\%. Segmentation examples are shown in \cref{fig:prostate}.

\textbf{Results on the Fundus dataset.}
In \cref{table: fundus}, we report the comparison results on the Fundus dataset. Among SSL methods, BCP stands out by creating challenging edges through copy-paste. DG methods surpass most of the SSL methods except BCP, likely due to the regular shapes of the optic cup and disc, making the task less challenging and allowing effective learning with even a few labeled data. RAM-DSIR, despite utilizing unlabeled data, underperforms compared to FedDG, possibly due to the consistency loss without ground truth guidance. Our method achieves the best Dice while CACPS achieves the best ASD. \cref{fig:fundus} displays the visualization results.

\input{tbl/mnms}
\begin{figure}[t]
   \centering
   \includegraphics[width=\linewidth]{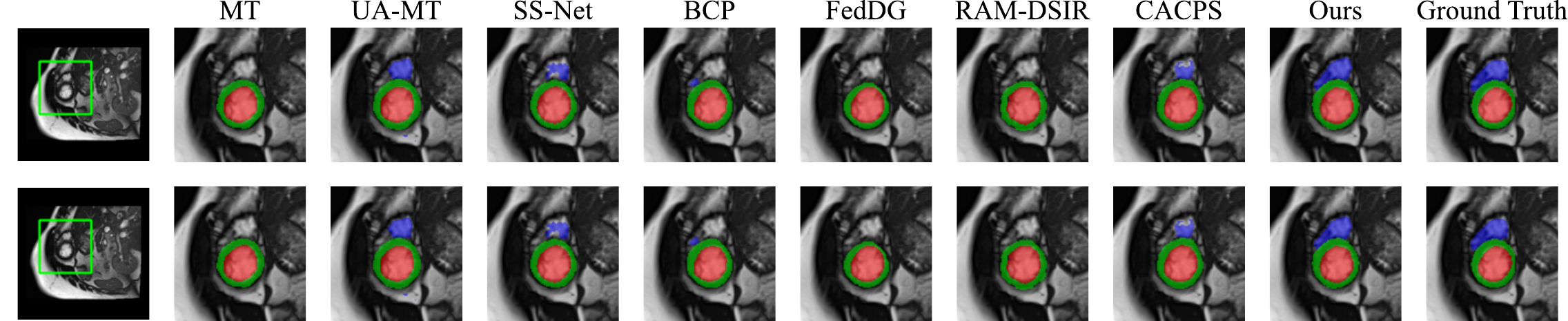}
   \caption{Visualization results from the M\&Ms dataset.}
   \label{fig:mnms}
\end{figure}

\textbf{Results on the M\&Ms dataset.} 
We conduct experiments on the M\&Ms dataset using 2\% and 5\% split ratios following \cite{yao2022cacps}. MT and UA-MT fail to generalize on the unseen domain on 2\% labeled data. 
Performance improves with more labeled samples for all methods, underscoring the importance of accurate supervision and better pseudo-label quality.
Our method outperforms all others on averaged metrics for both data splits.
Quantitative and visualized results are presented in \cref{table: mnms} and \cref{fig:mnms} respectively.

\subsection{Ablation Studies}
\label{sec:ablation}

\textbf{Effectiveness of each component.}
We perform an ablation study to evaluate the effectiveness of each component on the Fundus dataset. It can be observed from \cref{table: ablation} that the statistics-individual and aggregated branches (SIAB) improve the baseline method across all four domains. Additionally, the incorporation of histogram matching-based style augmentation (Style) contributes to enhanced generalization. Further improvements are observed with perturbations at the feature level through the random BN selection strategy (RandBN). Combining these three components forms the final framework of our method.

\textbf{Normalization statistics in statistics-aggregated branch.}
In \cref{table: ablation_inv}, we present experiment results about the choice of statistics used for normalization in SAB. Performance when solely using instance statistics for normalization is slightly inferior to using batch statistics. We believe this could be attributed to a small amount of task-relevant information contained within the entire batch, and normalizing with instance statistics dilutes such information. The integration of batch and instance statistics combines the individual strengths of both.

\begin{table}[tb!]
\parbox{.43\linewidth}{
\setlength\tabcolsep{1mm}
    \centering
    \small
    \caption{Ablation study (Dice \%) on the Fundus dataset.}
    \label{table: ablation}
    \resizebox{\linewidth}{!}{
    \begin{tabular}{ccc | ccccc}
        \toprule
        SIAB & Style & RandBN  & A & B & C & D & Avg. \\
        \midrule
         & & & 80.60 & 79.79 & 84.90 & 87.84 & 83.28 \\
        \checkmark  & & & 82.65 & 81.53 & 85.57 & 88.49 & 84.56 \\
         & \checkmark & & 81.95 & 78.53 & 85.92 & 89.12 & 83.88 \\
        \checkmark  & \checkmark & & 83.37 & 80.80 & 87.97 & 88.75 & 85.22 \\
        \midrule
        \checkmark & \checkmark & \checkmark & 83.95 & 80.93 & 87.30 & 90.24 & \textbf{85.60} \\
        \bottomrule
    \end{tabular}
    }
    \bigskip
    \setlength\tabcolsep{1.5mm}
    \centering
    \small
    \caption{Comparison (Dice \%) of different statistics for normalization in the SAB on the Fundus dataset.}
    \label{table: ablation_inv}
    \resizebox{\linewidth}{!}{
    \begin{tabular}{l | ccccc}
        \toprule
        Statistics & A & B & C & D & Avg. \\
        \midrule
        batch    & 81.73 & 80.09 & 84.50 & 87.69 & 83.50 \\
        instance & 81.37 & 79.70 & 84.42 & 87.77 & 83.31 \\
        \midrule
        batch \& instance & 82.65 & 81.53 & 85.57 & 88.49 & \textbf{84.56} \\
        \bottomrule
    \end{tabular}
    }
    \bigskip
    \setlength\tabcolsep{1.5mm}
    \caption{Comparison of different weights for RandBN on the Fundus dataset.}
    \label{table: wr}
    \resizebox{\linewidth}{!}{
    \begin{tabular}{c | cccccc}
        \toprule
        $\lambda_r$ & 0 & 0.05 & 0.1 & 0.2 & 0.5 & 1.0 \\
        \midrule
        Dice (\%) & 85.22 & 85.35 & 85.39 & \textbf{85.60} & 85.57 & 85.50 \\
        \bottomrule
    \end{tabular}}
}
\hfill
\parbox{.51\linewidth}{
\setlength\tabcolsep{1.7mm}
    \centering
    \small
    \caption{Comparison (Dice \%) of different image-level consistency regularization on the Fundus dataset.}
    \label{table: ablation_aug}
    \resizebox{\linewidth}{!}{
    \begin{tabular}{cc | ccccc}
        \toprule
        Aug1 & Aug2 & A & B & C & D & Avg. \\
        \midrule
        strong & -      & 82.42 & 80.98 & 84.46 & 89.06 & 84.23 \\
        style  & -      & 82.11 & 78.94 & 85.02 & 89.28 & 83.84 \\
        strong & strong & 83.27 & 80.75 & 85.37 & 89.66 & 84.76 \\
        \midrule
        strong & style  & 83.95 & 80.93 & 87.30 & 90.24 & \textbf{85.60} \\
        \bottomrule
    \end{tabular}}
    \bigskip
    \caption{Comparison of different probability $p$ for RandBN on the Fundus dataset.}
    \label{table: p}
    \resizebox{\linewidth}{!}{
    \begin{tabular}{c | cccccc}
        \toprule
        $p$ & 0 & 0.2 & 0.4 & 0.6 & 0.8 & 1.0 \\
        \midrule
        Dice (\%) & 85.18 & 85.42 & 85.44 & 85.48 & 85.60 & \textbf{85.63} \\
        \bottomrule
    \end{tabular}
    }
    \bigskip
    \setlength\tabcolsep{1.5mm}
    \caption{Runtime and learnable parameter comparison.}
    \label{table : eff}
    \resizebox{\linewidth}{!}{
    \begin{tabular}{l | cccc | c}
        \toprule
        Prostate & SS-Net & BCP & FedDG & CACPS & Ours \\
        \midrule
        Runtime       & 2.71h & 1.16h & 4.16h & 3.31h & 4.02h \\
        \#Par (Train) & 1.83M & 1.81M & 3.18M & 3.63M & 1.83M \\
        \#Par (Test)  & 1.81M & 1.81M & 3.18M & 1.81M & 1.81M \\
        \bottomrule
    \end{tabular}
    }
}
\end{table}

\textbf{Complementarity between the strong and style augmentation.}
We evaluate the effectiveness of distinct consistency regularization strategies, as summarized in \cref{table: ablation_aug}. Applying both the strong and style augmentation for consistency learning surpasses each individual stream. We also assess the performance by replacing style augmentation with another strong augmentation, resulting in lower outcomes. This suggests that the performance enhancement is not solely due to duplicating the augmentation stream but, more significantly, to the complementary relationship between strong and style augmentation.

\textbf{Balance of different consistency regularization.}
We apply the same weighting parameter ($\lambda_h = 1.0$) for weak-to-style consistency regularization as we do for weak-to-strong regularization, considering that both perturbations are performed at the image level.
We experiment to examine the effect of feature-level consistency regularization across varying values of $\lambda_r$. Table \ref{table: wr} shows that the averaged Dice score reaches its peak at 85.60\% with $\lambda_r$ set to 0.2.
However, further increasing $\lambda_r$ may result in a slight performance decrease.

\textbf{Intensity of RandBN perturbation.}
We experiment with the sensitivity of our method to variations in feature-level perturbation intensity. Results in Table \ref{table: p} indicate that, for all $p>0$, performance improves compared to the case when $p=0$, which represents vanilla consistency regularization without perturbation. This suggests that the effectiveness of RandBN perturbation is not trivially due to the introduction of an extra stream of regularization. As perturbation intensity increases, the model's generalization performance improves.

\textbf{Effect of the number of medical institutions.}
To investigate how the number of medical institutions affects the overall performance, we experiment on the Prostate dataset with fewer source domains and more target domains in \cref{table: source_domains}.
Despite performance drop with fewer source domains, our method outperforms CACPS more significantly than when using all 5 domains.

\begin{table}[t]
    \setlength\tabcolsep{0.9mm}
    \centering
    \small
    \caption{Experiments with fewer medical institutions on the Prostate dataset.}
    \label{table: source_domains}
    \resizebox{\linewidth}{!}{
    \begin{tabular}{l|ccc|ccc|ccc|cccc|cccc|c}
        \toprule
        Source & \multicolumn{3}{c|}{A D E} & \multicolumn{3}{c|}{A E F} & \multicolumn{3}{c|}{C D E} & \multicolumn{4}{c|}{A B} & \multicolumn{4}{c|}{B F} & \multirow{2}{*}{Avg.} \\
        \cmidrule{1-18} 
        Target & B & C & F & B & C & D & A & B & F & C & D & E & F & A & C & D & E & \\
        \midrule
        CACPS & 86.4 & 82.3 & 85.8 & 88.0 & 76.4 & 89.1 & 80.1 & 81.3 & \textbf{87.2} & 83.5 & \textbf{90.1} & 24.7 & 82.2 & 80.2 & 49.7 & 82.4 & 46.4 & 76.22 \\
        Ours  & \textbf{89.6} & \textbf{83.5} & \textbf{89.6} & \textbf{90.7} & \textbf{80.9} & \textbf{89.9} & \textbf{89.5} & \textbf{86.2} & 84.5 & \textbf{84.7} & 89.2 & \textbf{85.3} & \textbf{86.8} & \textbf{86.7} & \textbf{74.7} & \textbf{85.8} & \textbf{76.0} & \textbf{85.51} \\
        \bottomrule
    \end{tabular}
    }
\end{table}

\textbf{Discussion on runtime and parameter efficiency.}
We compare the training time and learnable parameter size in \cref{table : eff}.
Despite longer training time due to our multi-branch structure, we believe the cost is acceptable.
The quantity of learnable parameters in our method is also relatively low.

Due to the length limitation, please refer to supplementary materials for additional experiments on the ACDC\cite{bernard2018acdc} dataset and other hyperparameters.

%% file: tbl/prostate.tex
\begin{table*}[tb!]
\setlength\tabcolsep{0.9mm}
    \centering
    \footnotesize
    \caption{Performance comparisons with the state-of-the-art methods on the Prostate dataset.
    The best performance is marked as \textbf{bold}.
    }
    \label{table: prostate}
    \resizebox{\linewidth}{!}{
    \begin{tabular}{ll | ccccccc | ccccccc}
        \toprule
        \multicolumn{2}{c|}{\multirow{2}{*}{\textbf{Prostate}}} 
                & \multicolumn{7}{c|}{\textbf{Dice (\%)}}
                & \multicolumn{7}{c}{\textbf{ASD (voxel)}} \\
        \cmidrule{3-16} &   & A & B & C & D & E & F & Avg. & A & B & C & D & E & F & Avg.  \\
        \midrule
        \multicolumn{16}{c}{30\% Labeled}\\
        \midrule
        MT \cite{tarvainen2017meanteacher} & \pub{NIPS'17} & 79.20 & 57.20 & 47.03 & 86.51 & 12.81 & 84.59 & 61.23 & 1.17 & 1.59 & 6.73 & 3.81 & 43.65 & 0.84 & 9.63 \\
        UA-MT \cite{yu2019uncertainty} & \pub{MICCAI'19} & 83.14 & 64.80 & 68.97 & 87.64 & 26.57 & 85.47 & 69.43 & 0.99 & 1.40 & 8.38 & 1.28 & 20.94 & 0.97 & 5.66 \\
        SS-Net \cite{wu2022ssnet} & \pub{MICCAI'22} & 88.49 & 86.27 & 78.76 & 89.75 & 53.27 & 87.85 & 80.73 & 0.76 & 0.87 & 7.39 & 1.39 & 4.52 & 3.52 & 3.08 \\
        BCP \cite{bai2023bidirectional} & \pub{CVPR'23} & 88.38 & 87.11 & 64.01 & 89.59 & 69.12 & 84.89 & 80.52 & 3.08 & 0.74 & 19.29 & 1.13 & 5.94 & 11.09 & 6.88 \\
        CauSSL \cite{miao2023caussl} & \pub{ICCV'23} & 83.67 & 64.34 & 48.97 & 84.37 & 29.11 & 70.27 & 63.46 & 3.55 & 2.73 & 18.29 & 2.15 & 18.15 & 1.57 & 7.74 \\
        \midrule
        FedDG \cite{liu2021feddg} & \pub{CVPR'21} & 83.12 & 71.15 & 72.62 & 84.67 & 68.45 & 88.53 & 78.09 & 7.34 & 3.74 & 11.01 & 4.04 & 5.26 & 2.18 & 5.60 \\
        RAM-DSIR \cite{zhou2022ramdsir}	& \pub{ECCV'22} & 83.16 & 83.18 & 67.76 & 88.11 & 79.34 & 73.20 & 79.12 & 1.55 & 0.84 & 6.21 & 0.98 & \textbf{3.36} & 1.37 & 2.39 \\
        \midrule
        CACPS \cite{yao2022cacps} & \pub{AAAI'22} & \textbf{90.96}\stdev{2.96} & 88.18\stdev{4.60} & 79.20\stdev{14.3} & 89.16\stdev{3.46} & 81.12\stdev{11.7} & \textbf{90.19}\stdev{3.02} & 86.47 & 1.47\stdev{1.58} & 0.66\stdev{0.33} & 3.65\stdev{2.30} & 1.73\stdev{2.21} & 4.21\stdev{3.06} & 1.01\stdev{1.09} & 2.12 \\ 
        \midrule
        Ours & & 90.58\stdev{2.47} & \textbf{90.12}\stdev{2.95} & \textbf{85.72}\stdev{7.11} & \textbf{89.87}\stdev{2.90} & \textbf{86.06}\stdev{4.52} & 90.01\stdev{2.77} & \textbf{88.73} & \textbf{0.74}\stdev{0.46} & \textbf{0.57}\stdev{0.23} & \textbf{3.14}\stdev{2.35} & \textbf{0.72}\stdev{0.33} & 5.81\stdev{5.14} & \textbf{0.63}\stdev{0.23} & \textbf{1.93} \\
        \bottomrule
    \end{tabular}
    }
\end{table*}

%% file: tbl/fundus.tex
\begin{table*}[tb!]
\setlength\tabcolsep{0.8mm}
    \centering
    \footnotesize
    \caption{Performance (OC\,/\,OD) comparisons with the state-of-the-art methods on the Fundus dataset.
    The best performance is marked as \textbf{bold}.
    }
    \label{table: fundus}
    \resizebox{\linewidth}{!}{
    \begin{tabular}{ll | ccccc | ccccc}
        \toprule
        \multicolumn{2}{c|}{\multirow{2}{*}{\textbf{Fundus}}} 
                & \multicolumn{5}{c|}{\textbf{Dice (\%)}}
                & \multicolumn{5}{c}{\textbf{ASD (pixel)}} \\
        \cmidrule{3-12} &   & A & B & C & D & Avg. & A & B & C & D & Avg.  \\
        \midrule
        \multicolumn{12}{c}{30\% Labeled} \\
        \midrule
        MT \cite{tarvainen2017meanteacher} & \pub{NIPS'17} & 59.50\,/\,82.25 & 60.69\,/\,80.02 & 76.77\,/\,87.77 & 63.30\,/\,75.55 & 73.23 & 11.31\,/\,0.04 & 4.09\,/\,2.98 & 3.36\,/\,2.28 & 91.29\,/\,73.06 & 23.55 \\
        UA-MT \cite{yu2019uncertainty} & \pub{MICCAI'19} & 64.57\,/\,87.36 & 56.29\,/\,77.18 & 75.97\,/\,87.44 & 64.10\,/\,66.97 & 72.49 & 9.34\,/\,0.06 & 7.22\,/\,5.35 & 2.98\,/\,1.48 & 56.05\,/\,66.51 & 18.62 \\
        SS-Net \cite{wu2022ssnet} & \pub{MICCAI'22} & 71.64\,/\,92.37 & 57.84\,/\,69.03 & 80.81\,/\,90.75 & 63.65\,/\,65.12 & 73.90 & 6.12\,/\,0.07 & \textbf{2.60}\,/\,\textbf{1.60} & 1.65\,/\,0.30 & 47.83\,/\,66.92 & 15.89 \\
        BCP \cite{bai2023bidirectional} & \pub{CVPR'23} & 77.83\,/\,\textbf{95.14} & 64.22\,/\,74.58 & 81.61\,/\,91.04 & 80.25\,/\,90.86 & 81.94 & 4.43\,/\,0.21 & 4.46\,/\,3.27 & 2.57\,/\,0.65 & 2.11\,/\,1.46 & 2.39 \\
        CauSSL \cite{miao2023caussl} & \pub{ICCV'23} & 65.05\,/\,90.13 & 60.89\,/\,76.96 & 74.30\,/\,86.64 & 56.94\,/\,69.29 & 72.53 & 9.48\,/\,1.04 & 6.54\,/\,4.33 & 2.92\,/\,0.99 & 47.67\,/\,46.52 & 14.94 \\
        \midrule
        FedDG \cite{liu2021feddg} & \pub{CVPR'21} & \textbf{80.46}\,/\,93.76 & 66.60\,/\,84.61 & 82.52\,/\,90.90 & 79.62\,/\,90.50 & 83.62 & \textbf{3.61}\,/\,0.29 & 8.48\,/\,5.75 & 1.34\,/\,0.23 & 4.60\,/\,6.09 & 3.80 \\
        RAM-DSIR \cite{zhou2022ramdsir}	& \pub{ECCV'22} & 74.65\,/\,94.38 & 67.64\,/\,83.92 & 80.22\,/\,89.69 & 63.68\,/\,89.64 & 80.48 & 5.46\,/\,0.04 & 8.84\,/\,6.42 & \textbf{1.19}\,/\,\textbf{0.02} & 1.74\,/\,1.42 & 3.14 \\
        \midrule
        CACPS \cite{yao2022cacps} & \pub{AAAI'22} & 73.03\,/\,93.39 & 69.81\,/\,84.53 & 82.62\,/\,91.52 & 84.95\,/\,93.10 & 84.12 & 5.54\,/\,0.04 & 4.36\,/\,2.13 & 1.28\,/\,0.03 & 1.67\,/\,0.85 & \textbf{1.99} \\ 
        \midrule
        Ours & & 74.05\,/\,93.85 & \textbf{73.53}\,/\,\textbf{88.32} & \textbf{82.85}\,/\,\textbf{91.75} & \textbf{86.68}\,/\,\textbf{93.79} & \textbf{85.60} & 5.99\,/\,\textbf{0.04} & 4.21\,/\,2.70 & 1.25\,/\,0.07 & \textbf{1.42}\,/\,\textbf{0.79} & 2.06 \\
        \bottomrule
    \end{tabular}
    }
\end{table*}

%% file: tbl/mnms.tex
\begin{table*}[tb!]
\setlength\tabcolsep{1.3mm}
    \centering
    \footnotesize
    \caption{Performance comparisons with the state-of-the-art methods on the M\&Ms dataset.
    The best performance is marked as \textbf{bold} (same for the following tables).
    }
    \label{table: mnms}
    \resizebox{\linewidth}{!}{
    \begin{tabular}{ll | ccccc | ccccc}
        \toprule
        \multicolumn{2}{c|}{\multirow{2}{*}{\textbf{M\&Ms}}} 
                & \multicolumn{5}{c|}{\textbf{Dice (\%)}}
                & \multicolumn{5}{c}{\textbf{ASD (voxel)}} \\
        \cmidrule{3-12} &   & A & B & C & D & Avg. & A & B & C & D & Avg.  \\
        \midrule
        \multicolumn{12}{c}{2\% Labeled} \\
        \midrule
        SS-Net \cite{wu2022ssnet} & \pub{MICCAI'22} & 52.71 & 70.63 & 71.47 & 77.72 & 68.13 & 2.53 & 0.66 & 1.40 & 0.70 & 1.32 \\
        BCP \cite{bai2023bidirectional} & \pub{CVPR'23} & 47.62 & 83.57 & 80.00 & 81.45 & 73.16 &	1.83 & 1.54 & 1.97 & 2.66 & 2.00 \\
        FedDG \cite{liu2021feddg} & \pub{CVPR'21} & 60.04 & 65.40 & 67.46 & 61.61 & 63.63 & 4.07 & 3.71 & 7.41 & 6.89 & 5.52 \\
        RAM-DSIR \cite{zhou2022ramdsir}	& \pub{ECCV'22} & 24.13 & 68.43 & 52.16 & 46.41 & 47.78 &	7.59 & 1.97 & 6.52 & 1.96 & 4.51 \\
        CACPS \cite{yao2022cacps} & \pub{AAAI'22} & 82.35\stdev{6.24} & 82.84\stdev{7.59} & \textbf{86.31}\stdev{5.47} & 86.58\stdev{4.78} & 84.52 & 1.32\stdev{0.41} & 1.13\stdev{0.87} & 1.53\stdev{1.94} & 0.73\stdev{0.66} & 1.18 \\ 
        \midrule
        Ours & & \textbf{84.92}\stdev{4.19} & \textbf{86.80}\stdev{3.78} & 84.17\stdev{5.18} & \textbf{86.83}\stdev{2.07} & \textbf{85.68} & \textbf{0.54}\stdev{0.28} & \textbf{0.42}\stdev{0.23} & \textbf{0.94}\stdev{1.80} & \textbf{0.47}\stdev{0.21} & \textbf{0.59} \\
        \midrule
        \midrule
        \multicolumn{12}{c}{5\% Labeled} \\
        \midrule
        MT \cite{tarvainen2017meanteacher} & \pub{NIPS'17} & 30.59 & 66.46 & 70.61 & 61.88 & 57.39 & 13.37 & 1.60 & 7.41 & 1.82 & 6.05 \\
        UA-MT \cite{yu2019uncertainty} & \pub{MICCAI'19} & 51.76 & 45.98 & 28.43 & 48.78 & 43.73 & 10.80 & 3.41 & 10.37 & 5.52 & 7.53 \\
        SS-Net \cite{wu2022ssnet} & \pub{MICCAI'22} & 67.66 & 86.38 & 82.70 & 83.85 & 80.15 & 1.89 & 0.90 & 2.97 & 1.03 & 1.70 \\
        BCP \cite{bai2023bidirectional} & \pub{CVPR'23} & 70.15 & 87.26 & 82.71 & 82.53 & 80.66 & 2.16 & 1.32 & 3.61 & 1.29 & 2.09 \\
        FedDG \cite{liu2021feddg} & \pub{CVPR'21} & 75.44 & 83.76 & 81.07 & 76.89 & 79.29 & 1.70 & 1.92 & 2.57 & 4.93 & 2.78 \\
        RAM-DSIR \cite{zhou2022ramdsir}	& \pub{ECCV'22} & 53.93 & 81.66 & 74.22 & 82.97 & 73.19 & 2.67 & 0.78 & 0.66 & 1.22 & 1.33 \\
        CACPS \cite{yao2022cacps} & \pub{AAAI'22} & 83.30\stdev{5.83} & 85.04\stdev{6.49} & \textbf{87.14}\stdev{4.74} & 87.38\stdev{4.49} & 85.81 & 1.01\stdev{0.34} & 0.61\stdev{0.75} & 1.26\stdev{1.86} & \textbf{0.36}\stdev{0.56} & 0.81 \\ 
        \midrule
        Ours & & \textbf{85.27}\stdev{3.54} & \textbf{87.84}\stdev{3.07} & 86.03\stdev{4.20} & \textbf{87.57}\stdev{2.25} & \textbf{86.68} & \textbf{0.51}\stdev{0.31} & \textbf{0.40}\stdev{0.16} & \textbf{0.85}\stdev{1.64} & 0.37\stdev{0.08} & \textbf{0.53}  \\
        \bottomrule
    \end{tabular}
    }
\end{table*}

%% file: sec/5_cl.tex
\section{Conclusion}
\label{sec:cl}

In this paper, we propose a novel framework tackling semi-supervised domain generalization in medical image segmentation. Our statistics-individual and statistics-aggregated branches effectively mitigate the adverse effects of statistics differences on pseudo-labels, while simultaneously leveraging them to enhance generalization performance. 
Additionally, we introduce perturbations at both image and feature levels, expanding the consistency regularization for a broader perturbation exploration. Evaluation on multiple benchmarks demonstrates its superiority over other SOTA methods.

%% file: sec/X_suppl.tex
\appendix
\onecolumn
\begin{center}
  \Large\bfseries The Devil is in the Statistics: Mitigating and Exploiting Statistics Difference for Generalizable Semi-supervised Medical Image Segmentation \\
  \large - Supplementary Material -
\end{center}

\renewcommand{\thesection}{\Alph{section}}
\setcounter{section}{0}

\section{Additional Implementation Details}

We implement the statistics-individual and statistics-aggregated branches in a non-intrusive manner. With just one line of code, we can replace the BN layers in most networks with statistics-individual and aggregated branches without the need to modify the source code defining the network, boosting efficiency while avoiding potential pitfalls linked to modifying the original code. Specifically, our process involves traversing and replacing BN modules. To enable the modules to accept additional parameters indicating domain identifiers, we leveraged PyTorch's forward hook mechanism.
In \cref{alg:unsupervised}, we provide the pseudocode for the consistency regularization on unlabeled data, incorporating the integration of the statistics-individual and aggregated branches.

For the learnable mixing coefficient $\alpha$ in \cref{eq:inv},
we utilize the sigmoid function to ensure its range stays within 0 to 1. Additionally, we employ a learning rate 10 times larger than other parameters with weight decay set to 0 for updating the mixing coefficient. The mixing coefficients on the Prostate, Fundus, and M\&Ms datasets are empirically initialized as 0.999, 0.5, and 0.999 respectively. The confidence threshold $\tau$ in \cref{eq:unlabeled-full} is consistently set to $0.95$ across all datasets.

\begin{algorithm}[t]
\caption{Pseudocode for unsupervised loss}
\label{alg:unsupervised}
\definecolor{codeblue}{rgb}{0.25,0.5,0.5}
\lstset{
  backgroundcolor=\color{white},
  basicstyle=\fontsize{7pt}{7pt}\ttfamily\selectfont,
  columns=fullflexible,
  breaklines=true,
  captionpos=b,
  commentstyle=\fontsize{7pt}{7pt}\color{codeblue},
  keywordstyle=\fontsize{7pt}{7pt}
}
\begin{lstlisting}[language=python]
# fs: student segmentation network
# ft: teacher segmentation network
# aug_w/aug_s/aug_h: weak/strong/hist-match augmentation
# k: number of source domains
# thresh: confidence threshold
# lambda_h/lambda_r: weights of consistency

# set up statistics-individual and aggregated branches
# the last domain denotes the statistics-aggregated branch
\end{lstlisting}
\vspace{-\baselineskip}
\begin{lstlisting}[language=python,backgroundcolor=\color{lightgray}]
SIAB.convert(fs, n_domains=k + 1)  # single line for model conversion
SIAB.convert(ft, n_domains=k + 1)
\end{lstlisting}
\begin{lstlisting}[language=python]
for u, domain_u in loader_u:
    # domain_u contains domain labels for each sample
    u_w = aug_w(u)
    u_s = aug_s(u_w)
    u_h = aug_h(u_w)

    # obtain predictions with SIAB
    p_w_ind = ft(u_w, domain_id=domain_u)
    p_w_agg = ft(u_w, domain_id=k + 1)
    p_w = (p_w_ind + p_w_agg) / 2
    p_s = fs(u_s, domain_id=k + 1)
    p_h = fs(u_h, domain_id=k + 1)
    with SIAB.stop_bn_gradient(fs):
        p_r = fs(u_w, domain_id=domain_u, rand=True)

    # pseudo label and thresholding
    conf, pl = p_w.detach().softmax(dim=1).max(dim=1)
    mask = (conf > thresh).float()

    # loss
    criterion = nn.CrossEntropyLoss(reduction='none')
    loss_s = (criterion(p_s, pl) * mask).mean()
    loss_h = (criterion(p_h, pl) * mask).mean()
    loss_r = (criterion(p_r, pl) * mask).mean()
    loss_u = loss_s + lambda_h * loss_h + lambda_r + loss_r
\end{lstlisting}
\end{algorithm}

\section{Additional Experiments}

\noindent \textbf{Experiments on typical semi-supervised medical image segmentation tasks.}
Since our method can also be applied to semi-supervised tasks, we experiment on the ACDC \cite{bernard2018acdc} dataset and compare the performance against the state-of-the-art semi-supervised medical image segmentation methods.
While specifically designed for semi-supervised domain generalization with data from multiple medical institutions, our method attains competitive results on typical semi-supervised tasks.

\begin{table}[h]
\setlength\tabcolsep{1.9mm}
    \footnotesize
    \caption{Comparisons with the state-of-the-art methods on the ACDC dataset.}
    \centering
    \begin{tabular}{l | cccc | c}
        \toprule
        ACDC (Dice, \%) & UA-MT & SS-Net & BCP & CauSSL & Ours \\
        \midrule
        3 cases (5\%)  & 46.04 & 65.83 & 87.59 & - & 86.72 \\
        7 cases (10\%) & 81.65 & 86.78 & 88.84 & 89.66 & 88.67 \\
        \bottomrule
    \end{tabular}
    \label{tab:my_label}
\end{table}

\noindent \textbf{Experiments with reduced training iterations.}
We provide the performance of our method using different training time on the 30\% labeled Prostate dataset and compare it to the previous SOTA CACPS.
As shown below, our method consistently outperforms CACPS.
Notably, even with just 2 hours of training time, our method still exhibits a significant advantage of 1.5\%.

\begin{table}[h]
\setlength\tabcolsep{1.9mm}
    \footnotesize
    \caption{Results with reduced training iterations on the Prostate dataset.}
    \centering
    \begin{tabular}{l|llll}
        \toprule
        Prostate (30\%) & CACPS & Ours &  Ours & Ours \\
        \midrule
        Time (h)   & 3.31  & 2.01 & 3.22 & 4.02 \\
        Dice (\%)  & 86.47 &87.98 \textcolor{mygreen}{+\textbf{1.51}} &  88.58 \textcolor{mygreen}{+\textbf{2.11}} & 88.73 \textcolor{mygreen}{+\textbf{2.26}} \\
        \bottomrule
    \end{tabular}
\end{table}

\noindent \textbf{Effect of more weight factors.}
We implicitly assigned equal weights to the factors in the following equations:
\begin{itemize}
    \item $ \textbf{Unsupervised loss:}$ $\mathcal{L} = \mathcal{L}_{x} + \lambda_u \mathcal{L}_{u},$
    \item $ \textbf{Pseudo-label ensemble:}$ $p_w = t \cdot p_w^{\mathcal{IF}} + (1 - t) \cdot p_w^{\mathcal{AF}},$
    \item $ \textbf{Supervised loss:}$ $\mathcal{L}_x = \mathcal{L}_{\mathcal{IF}} + \lambda_{\mathcal{AF}} \mathcal{L}_{\mathcal{AF}},$
\end{itemize}
where $\mathcal{L}_{\mathcal{IF}}$ and $\mathcal{L}_{\mathcal{AF}}$ denote the supervised loss for statistics-individual and aggregated branches, respectively.
We empirically investigate the impact of these weight factors on the 30\% labeled Fundus dataset. The results are shown below with default weight factors marked in \textcolor{blue}{blue} and best performances in \textbf{bold}.
Our method demonstrates robustness against variation with different values of weighting factors. Setting the weights equally achieves near-optimal performance, with marginal gains observed upon further adjustment.

\begin{table}[h]
\setlength\tabcolsep{2.42mm}
    \centering
    \footnotesize
    \caption{Experiments on different weight factors on the Fundus dataset.}
    \begin{tabular}{c|cccccc}
        \toprule
        $\lambda_u$ & 0.3 & 0.5 & 0.7 & \textcolor{blue}{{1.0}} & 1.5 & 2.0  \\
        \midrule
        Dice (\%) & 85.27 & 85.63 & \textbf{85.74} & 85.60 & 85.39 & 85.44 \\
        \bottomrule
    \end{tabular}
\end{table}
\vspace{-2\baselineskip}
\begin{figure}[h]
\begin{subfigure}[t]{0.45\linewidth}
    \setlength\tabcolsep{1.5mm}
    \centering
    \footnotesize
    \begin{tabular}{c|ccc}
        \toprule
        $t$ & 0.25 & \textcolor{blue}{{0.5}} & 0.75  \\
        \midrule
        Dice (\%) & 85.31 & \textbf{85.60} & 85.34 \\
        \bottomrule
    \end{tabular}
\end{subfigure}%
\hfill
\begin{subfigure}[t]{0.52\linewidth}
    \setlength\tabcolsep{1.5mm}
    \centering
    \footnotesize
    \begin{tabular}{c|cccc}
        \toprule
        $\lambda_{\mathcal{AF}}$ & 0.5 & \textcolor{blue}{{1.0}} & 1.5 & 2.0  \\
        \midrule
        Dice (\%) & 84.97 & 85.60 & \textbf{85.69} & 85.34 \\
        \bottomrule
    \end{tabular}
\end{subfigure}%
\end{figure}

\noindent \textbf{Effect of confidence threshold.}
We evaluate the sensitivity of our method against varying confidence thresholds on the Fundus dataset as follows. The performance variation with different values of $\tau$ is minimal.

\begin{table}[h]
    \setlength\tabcolsep{2.8mm}
    \centering
    \footnotesize
    \caption{Experiments on confidence threshold on the Fundus dataset.}
    \begin{tabular}{c|ccccc}
        \toprule
        $\tau$ & 0.85 & 0.90 & 0.95 & 0.98 & 0.99 \\
        \midrule
        Dice (\%) & 85.30 & 85.39 & \textbf{85.60} & 85.46 & 85.34 \\
        \bottomrule
    \end{tabular}
\end{table}

\section{Limitations}

In the previous state-of-the-art method, CACPS~\cite{yao2022cacps}, for medical image segmentation in SSDG, domain labels were not utilized, while our approach introduces statistics-individual branches that necessitate the use of domain labels. We believe that obtaining domain labels is relatively straightforward. For instance, in collaborative training involving multiple medical institutions, using medical institution information as domain identifiers is a viable approach. Therefore, it is reasonable to utilize domain labels in SSDG, following the conventional practices in domain generalization.

%% file: main.bbl
\begin{thebibliography}{10}
\providecommand{\url}[1]{\texttt{#1}}
\providecommand{\urlprefix}{URL }
\providecommand{\doi}[1]{https://doi.org/#1}

\bibitem{bai2017semi}
Bai, W., Oktay, O., Sinclair, M., Suzuki, H., Rajchl, M., Tarroni, G., Glocker, B., King, A., Matthews, P.M., Rueckert, D.: Semi-supervised learning for network-based cardiac mr image segmentation. In: Medical Image Computing and Computer Assisted Intervention. pp. 253--260. Springer (2017)

\bibitem{bai2023bidirectional}
Bai, Y., Chen, D., Li, Q., Shen, W., Wang, Y.: Bidirectional copy-paste for semi-supervised medical image segmentation. In: Proceedings of the IEEE/CVF Conference on Computer Vision and Pattern Recognition. pp. 11514--11524 (2023)

\bibitem{bernard2018acdc}
Bernard, O., Lalande, A., Zotti, C., Cervenansky, F., Yang, X., Heng, P.A., Cetin, I., Lekadir, K., Camara, O., Ballester, M.A.G., et~al.: Deep learning techniques for automatic mri cardiac multi-structures segmentation and diagnosis: is the problem solved? IEEE Transactions on Medical Imaging  \textbf{37}(11),  2514--2525 (2018)

\bibitem{cai2023orthogonal}
Cai, H., Li, S., Qi, L., Yu, Q., Shi, Y., Gao, Y.: Orthogonal annotation benefits barely-supervised medical image segmentation. In: Proceedings of the IEEE/CVF Conference on Computer Vision and Pattern Recognition. pp. 3302--3311 (2023)

\bibitem{campello2021multi}
Campello, V.M., Gkontra, P., Izquierdo, C., Martin-Isla, C., Sojoudi, A., Full, P.M., Maier-Hein, K., Zhang, Y., He, Z., Ma, J., et~al.: Multi-centre, multi-vendor and multi-disease cardiac segmentation: the m\&ms challenge. IEEE Transactions on Medical Imaging  \textbf{40}(12),  3543--3554 (2021)

\bibitem{chen2021transunet}
Chen, J., Lu, Y., Yu, Q., Luo, X., Adeli, E., Wang, Y., Lu, L., Yuille, A.L., Zhou, Y.: Transunet: Transformers make strong encoders for medical image segmentation. arXiv preprint arXiv:2102.04306  (2021)

\bibitem{chen2021cps}
Chen, X., Yuan, Y., Zeng, G., Wang, J.: Semi-supervised semantic segmentation with cross pseudo supervision. In: Proceedings of the IEEE/CVF Conference on Computer Vision and Pattern Recognition. pp. 2613--2622 (2021)

\bibitem{cheplygina2019not}
Cheplygina, V., de~Bruijne, M., Pluim, J.P.: Not-so-supervised: a survey of semi-supervised, multi-instance, and transfer learning in medical image analysis. Medical Image Analysis  \textbf{54},  280--296 (2019)

\bibitem{dou2019domain}
Dou, Q., Coelho~de Castro, D., Kamnitsas, K., Glocker, B.: Domain generalization via model-agnostic learning of semantic features. Advances in Neural Information Processing Systems  \textbf{32} (2019)

\bibitem{guo2023domaindrop}
Guo, J., Qi, L., Shi, Y.: Domaindrop: Suppressing domain-sensitive channels for domain generalization. In: Proceedings of the IEEE/CVF International Conference on Computer Vision. pp. 19114--19124 (2023)

\bibitem{hu2022domain}
Hu, S., Liao, Z., Zhang, J., Xia, Y.: Domain and content adaptive convolution based multi-source domain generalization for medical image segmentation. IEEE Transactions on Medical Imaging  \textbf{42}(1),  233--244 (2022)

\bibitem{ioffe2015batch}
Ioffe, S., Szegedy, C.: Batch normalization: Accelerating deep network training by reducing internal covariate shift. In: International Conference on Machine Learning. pp. 448--456. pmlr (2015)

\bibitem{lee2013pseudo}
Lee, D.H., et~al.: Pseudo-label: The simple and efficient semi-supervised learning method for deep neural networks. In: Workshop on challenges in representation learning, ICML. vol.~3, p.~896. Atlanta (2013)

\bibitem{li2018learning}
Li, D., Yang, Y., Song, Y.Z., Hospedales, T.: Learning to generalize: Meta-learning for domain generalization. In: Proceedings of the AAAI conference on artificial intelligence. vol.~32 (2018)

\bibitem{li2020transformation}
Li, X., Yu, L., Chen, H., Fu, C.W., Xing, L., Heng, P.A.: Transformation-consistent self-ensembling model for semisupervised medical image segmentation. IEEE Transactions on Neural Networks and Learning Systems  \textbf{32}(2) (2020)

\bibitem{liu2021feddg}
Liu, Q., Chen, C., Qin, J., Dou, Q., Heng, P.A.: Feddg: Federated domain generalization on medical image segmentation via episodic learning in continuous frequency space. In: Proceedings of the IEEE/CVF Conference on Computer Vision and Pattern Recognition. pp. 1013--1023 (2021)

\bibitem{liu2020saml}
Liu, Q., Dou, Q., Heng, P.A.: Shape-aware meta-learning for generalizing prostate mri segmentation to unseen domains. In: Medical Image Computing and Computer Assisted Intervention. pp. 475--485. Springer (2020)

\bibitem{liu2021semimeta}
Liu, X., Thermos, S., O’Neil, A., Tsaftaris, S.A.: Semi-supervised meta-learning with disentanglement for domain-generalised medical image segmentation. In: Medical Image Computing and Computer Assisted Intervention. Springer (2021)

\bibitem{Loshchilov2019adamw}
Loshchilov, I., Hutter, F.: Decoupled weight decay regularization. In: International Conference on Learning Representations (2019)

\bibitem{luo2021semi}
Luo, X., Chen, J., Song, T., Wang, G.: Semi-supervised medical image segmentation through dual-task consistency. In: Proceedings of the AAAI conference on artificial intelligence. vol.~35, pp. 8801--8809 (2021)

\bibitem{lyu2022aadg}
Lyu, J., Zhang, Y., Huang, Y., Lin, L., Cheng, P., Tang, X.: Aadg: automatic augmentation for domain generalization on retinal image segmentation. IEEE Transactions on Medical Imaging  \textbf{41}(12),  3699--3711 (2022)

\bibitem{miao2023caussl}
Miao, J., Chen, C., Liu, F., Wei, H., Heng, P.A.: Caussl: Causality-inspired semi-supervised learning for medical image segmentation. In: Proceedings of the IEEE/CVF International Conference on Computer Vision. pp. 21426--21437 (2023)

\bibitem{milletari2016vnet}
Milletari, F., Navab, N., Ahmadi, S.A.: V-net: Fully convolutional neural networks for volumetric medical image segmentation. In: 2016 fourth international conference on 3D vision (3DV). pp. 565--571. IEEE (2016)

\bibitem{nam2018batch}
Nam, H., Kim, H.E.: Batch-instance normalization for adaptively style-invariant neural networks. Advances in Neural Information Processing Systems  \textbf{31} (2018)

\bibitem{pizer1987adaptive}
Pizer, S.M., Amburn, E.P., Austin, J.D., Cromartie, R., Geselowitz, A., Greer, T., ter Haar~Romeny, B., Zimmerman, J.B., Zuiderveld, K.: Adaptive histogram equalization and its variations. Computer vision, graphics, and image processing  \textbf{39}(3),  355--368 (1987)

\bibitem{ronneberger2015unet}
Ronneberger, O., Fischer, P., Brox, T.: U-net: Convolutional networks for biomedical image segmentation. In: Medical Image Computing and Computer Assisted Intervention. pp. 234--241. Springer (2015)

\bibitem{seo2020learning}
Seo, S., Suh, Y., Kim, D., Kim, G., Han, J., Han, B.: Learning to optimize domain specific normalization for domain generalization. In: Proceedings of the European Conference on Computer Vision. pp. 68--83. Springer (2020)

\bibitem{shi2021inconsistency}
Shi, Y., Zhang, J., Ling, T., Lu, J., Zheng, Y., Yu, Q., Qi, L., Gao, Y.: Inconsistency-aware uncertainty estimation for semi-supervised medical image segmentation. IEEE Transactions on Medical Imaging  \textbf{41}(3),  608--620 (2021)

\bibitem{sohn2020fixmatch}
Sohn, K., Berthelot, D., Carlini, N., Zhang, Z., Zhang, H., Raffel, C.A., Cubuk, E.D., Kurakin, A., Li, C.L.: Fixmatch: Simplifying semi-supervised learning with consistency and confidence. Advances in Neural Information Processing Systems  \textbf{33},  596--608 (2020)

\bibitem{su2023rethinking}
Su, Z., Yao, K., Yang, X., Huang, K., Wang, Q., Sun, J.: Rethinking data augmentation for single-source domain generalization in medical image segmentation. In: Proceedings of the AAAI Conference on Artificial Intelligence. vol.~37, pp. 2366--2374 (2023)

\bibitem{tajbakhsh2020embracing}
Tajbakhsh, N., Jeyaseelan, L., Li, Q., Chiang, J.N., Wu, Z., Ding, X.: Embracing imperfect datasets: A review of deep learning solutions for medical image segmentation. Medical Image Analysis  \textbf{63},  101693 (2020)

\bibitem{tarvainen2017meanteacher}
Tarvainen, A., Valpola, H.: Mean teachers are better role models: Weight-averaged consistency targets improve semi-supervised deep learning results. Advances in Neural Information Processing Systems  \textbf{30} (2017)

\bibitem{wang2022generalizing}
Wang, J., Lan, C., Liu, C., Ouyang, Y., Qin, T., Lu, W., Chen, Y., Zeng, W., Yu, P.: Generalizing to unseen domains: A survey on domain generalization. IEEE Transactions on Knowledge and Data Engineering  (2022)

\bibitem{wang2020dofe}
Wang, S., Yu, L., Li, K., Yang, X., Fu, C.W., Heng, P.A.: Dofe: Domain-oriented feature embedding for generalizable fundus image segmentation on unseen datasets. IEEE Transactions on Medical Imaging  \textbf{39}(12),  4237--4248 (2020)

\bibitem{wang2023generalizable}
Wang, X., Zhang, J., Qi, L., Shi, Y.: Generalizable decision boundaries: Dualistic meta-learning for open set domain generalization. In: Proceedings of the IEEE/CVF International Conference on Computer Vision. pp. 11564--11573 (2023)

\bibitem{wang2023mcf}
Wang, Y., Xiao, B., Bi, X., Li, W., Gao, X.: Mcf: Mutual correction framework for semi-supervised medical image segmentation. In: Proceedings of the IEEE/CVF Conference on Computer Vision and Pattern Recognition. pp. 15651--15660 (2023)

\bibitem{wu2022ssnet}
Wu, Y., Wu, Z., Wu, Q., Ge, Z., Cai, J.: Exploring smoothness and class-separation for semi-supervised medical image segmentation. In: Medical Image Computing and Computer Assisted Intervention. pp. 34--43. Springer (2022)

\bibitem{wu2021semi}
Wu, Y., Xu, M., Ge, Z., Cai, J., Zhang, L.: Semi-supervised left atrium segmentation with mutual consistency training. In: Medical Image Computing and Computer Assisted Intervention. pp. 297--306. Springer (2021)

\bibitem{yang2023revisiting}
Yang, L., Qi, L., Feng, L., Zhang, W., Shi, Y.: Revisiting weak-to-strong consistency in semi-supervised semantic segmentation. In: Proceedings of the IEEE/CVF Conference on Computer Vision and Pattern Recognition. pp. 7236--7246 (2023)

\bibitem{yao2022cacps}
Yao, H., Hu, X., Li, X.: Enhancing pseudo label quality for semi-supervised domain-generalized medical image segmentation. In: Proceedings of the AAAI Conference on Artificial Intelligence. vol.~36, pp. 3099--3107 (2022)

\bibitem{yu2019uncertainty}
Yu, L., Wang, S., Li, X., Fu, C.W., Heng, P.A.: Uncertainty-aware self-ensembling model for semi-supervised 3d left atrium segmentation. In: Medical Image Computing and Computer Assisted Intervention. pp. 605--613. Springer (2019)

\bibitem{yuan2021simple}
Yuan, J., Liu, Y., Shen, C., Wang, Z., Li, H.: A simple baseline for semi-supervised semantic segmentation with strong data augmentation. In: Proceedings of the IEEE/CVF International Conference on Computer Vision. pp. 8229--8238 (2021)

\bibitem{zhou2021stylematch}
Zhou, K., Loy, C.C., Liu, Z.: Semi-supervised domain generalization with stochastic stylematch. International Journal of Computer Vision pp. 1--11 (2023)

\bibitem{zhou2022ramdsir}
Zhou, Z., Qi, L., Shi, Y.: Generalizable medical image segmentation via random amplitude mixup and domain-specific image restoration. In: Proceedings of the European Conference on Computer Vision. pp. 420--436. Springer (2022)

\end{thebibliography}
